\documentclass[preprint,12pt,authoryear]{elsarticle}

\usepackage{amssymb}

\usepackage{amsmath}
\usepackage{amsfonts}

\usepackage[onehalfspacing]{setspace}
\usepackage{booktabs}
\usepackage{amsmath}
\usepackage{dsfont}
\usepackage{booktabs}
\usepackage{mathrsfs}
\usepackage{xcolor}
\usepackage{hyperref}
\usepackage{cleveref}

\newcommand*{\expect}[1]{\mathrm{E}[#1]}
\newcommand*{\expecttwo}[1]{\mathrm{E}^2[#1]}
\newcommand*{\variance}[1]{\mathrm{V}[#1]}
\newcommand*{\stdev}[1]{\mathrm{S}[#1]}
\newcommand*{\diff}{\, \mathrm{d}}
\newcommand*{\euler}[1]{\text{B}(#1)}
\newcommand*{\classorder}[1]{\mathcal{O}(\mathcal{C}_#1)}
\newcommand*{\class}[1]{\mathcal{C}_#1}

\journal{Expert Systems with Applications}

\begin{document}

\begin{frontmatter}



\title{Improving the classification of extreme classes by means of loss regularisation and generalised beta distributions}


\author[uco]{V\'ictor Manuel Vargas\corref{cor1}}
\ead{vvargas@uco.es}


\address[ayrna]{Department of Computer Science and Numerical Analysis, University of C\'ordoba, Rabanales Campus, Albert Einstein building, 14014 C\'ordoba, Spain
				}

\author[uco]{Pedro Antonio Guti\'errez}
\ead{pagutierrez@uco.es}
\author[uco]{Javier Barbero G\'omez}
\ead{jbarbero@uco.es}
\author[uco]{C\'esar Herv\'as-Mart\'inez}
\ead{chervas@uco.es}

\cortext[cor1]{Corresponding author}

\begin{abstract}
An ordinal classification problem is one in which the target variable takes values on an ordinal scale. Nowadays, there are many of these problems associated with real-world tasks where it is crucial to accurately classify the extreme classes of the ordinal structure. In this work, we propose a unimodal regularisation approach that can be applied to any loss function to improve the classification performance of the first and last classes while maintaining good performance for the remainder. The proposed methodology is tested on six datasets with different numbers of classes, and compared with other unimodal regularisation methods in the literature. In addition, performance in the extreme classes is compared using a new metric that takes into account their sensitivities. Experimental results and statistical analysis show that the proposed methodology obtains a superior average performance considering different metrics. The results for the proposed metric show that the generalised beta distribution generally improves classification performance in the extreme classes. At the same time, the other five nominal and ordinal metrics considered show that the overall performance is aligned with the performance of previous alternatives. 
\end{abstract}

\begin{keyword}
extreme classes \sep ordinal regression \sep convolutional neural networks \sep regularised loss \sep generalised beta distribution
\end{keyword}

\end{frontmatter}

\section{Introduction}
In the last decade, the use of machine learning and deep learning techniques to solve classification tasks has received an increasing interest in the literature, due to their multiple real-world applications in different areas such as industry \citep{bertolini2021machine,vargas2023deep,zhang2021adaptive,jimenez2020validation}, medicine \citep{houssein2021deep,saibene2021expert}, internet of things \citep{zhang2021big,klaib2021eye} or renewable energies \citep{gomez2022simultaneous}. Some of these problems have an implicit order in the categories to be predicted, in such a way that they can be naturally ordered following a scale determined by the real problem. Solving these kind of tasks is commonly known as ordinal classification or ordinal regression \citep{perez2014projection,riccardi2014cost}, given that they share some characteristics with both classification and regression problems. Therefore, solving an ordinal classification problem consists on predicting the correct category from a set of discrete categories that are arranged following an order determined by the real problem. In contrast to regression problems, the predicted class is not continuous and the distance between two adjacent classes does not have to be the same for all the categories.

In the machine learning context, a classification problem can be formally defined as the problem of predicting the label $y$ using an input vector $\mathbf{x}$, where $\mathbf{x} \in \mathcal{X} \subseteq \mathds{R}^d$ and $y \in \mathcal{Y}=\{\class{1}, \class{2}, ..., \class{J}\}$, where $d \in \mathds{N}^+$ and $J$ is the number of categories of the problem. To solve these problems, the objective is to find a function $r: \mathcal{X} \rightarrow \mathcal{Y}$ that predicts the category of any given sample using the input data. In the particular case of an ordinal classification problem, categories have an intrinsic order which can be defined by the following order constraint: $\class{1} \prec \class{2} \prec ... \prec \class{J}$. Although the order between categories is defined by the problem, it is common for the distance between these categories to be unknown and non-uniform. Thus, the order of a category is given by $\classorder{j} = j$, and the previous constraint can be also expressed as $\classorder{1} < \classorder{2} < ... < \classorder{J}$. Taking into account the specific characteristics of ordinal problems, different methodologies have recently been proposed in the literature to consider these elements and attempt to improve classification performance by reducing the magnitude of errors. This is achieved by avoiding mistakes in distant classes, which incur a high penalty in this type of problem. In \cite{singer2020weighted}, the authors proposed a weighted information-gain metric. They used this metric to generate decision trees for ordinal classification problems, resulting in improved performance compared to using a nominal metric. Furthermore, in \cite{singer2021classification}, the authors suggested using a decision-tree model with the aforementioned weighted information-gain metric to solve a medical problem involving the diagnosis of the severity of trachea stenosis from EEG signals. Additionally, in \cite{barbero2021ordinal}, the authors applied a different ordinal approach to assess neurological damage in Parkinson's disease patients. This approach involved decomposing the original ordinal problem into several binary problems, which could be solved separately or jointly. In \cite{tang2021comparative}, the authors proposed a methodology that overcomes one of the most common problems in ordinal classification: the lack of data. To do that, they add additional information to each sample apart from its label. Besides, in \cite{lazaro2023neural} the authors proposed an approach oriented to solve another common problem in ordinal tasks, which is their imbalance nature. To do that, they propose a loss function which is an estimate of the Bayesian classification cost. With the same concern, the authors of \cite{zhu2019minority} proposed a generation direction-aware synthetic minority oversampling technique to deal exclusively with imbalanced ordinal regression problems. Finally, in \cite{tang2022ordinal}, the authors proposed a distance metric learning method specifically designed to deal with the combination of absolute and frequentist relative information in ordinal classification problems.

In most nominal classification tasks, the importance of all the categories of the problem is the same. However, there are many ordinal classification tasks where the extreme classes (i.e., the first and the last classes) are more relevant than the others. Therefore, improving the sensitivity of these classes can be quite interesting for the specific real-world problem addressed. For example, in a 4-class bio-medical problem where the severity of a disease is predicted \citep{khan2020automatic,atwany2022deep}, and level 0 is related to a healthy patient while level 4 corresponds to the worst grade of the disease, correctly predicting these two classes can be very important. Also, in an Industry 4.0 quality control problem \citep{rosati2022novel,damacharla2021tlu} where any manufactured product is classified depending on its quality, accurately predicting the worst class can be crucial to remove products that do not satisfy the minimum quality requirements. In the same way, properly predicting the best class can also be very interesting to form a ``high-quality product range'' category and sell them separately.


Moreover, there are some problems where there is an important presence of noise in the target labels due to the intrinsic characteristics of the problem. In most cases, the samples have been labelled by an expert who may make misclassifications. This fact becomes more evident when the class labels of the problem have a natural order. In such cases, the expert who has labelled the data can classify two patterns that are very similar in two different adjacent classes. Even though the samples could be perfectly labelled, as in the case of classifying faces by age ranges, a model that classifies these samples using a standard $0/1$ label encoding can perform poorly on the border of two categories. To address this problem of noisy labels, label smoothing was introduced by  \cite{liu2020unimodal}. In this work, the authors proposed to train a neural network model using a regularised loss function that transforms the standard $0/1$ label encoding into a soft alternative that better represents the labels in an ordinal classification problem. Given that labeling errors usually occur in adjacent classes due to their similarity, they proposed using unimodal distributions to define the new label encoding. Furthermore, they compared the performance of Poisson, binomial, and exponential distributions when used to generate the new label encoding. It is worth noting that the Poisson distribution is a discrete distribution that only has one parameter, which directly determines both the mean and the variance of the distribution. For this reason, it is not possible to find a value that places the mean in the middle of the class interval for every class in a problem with $J$ classes while keeping the variance low, which is desirable. Therefore, in the same work, the authors proposed the binomial distribution, which uses different expressions for the mean and the variance, making it easier to adjust them separately. However, for some classes, it is still not possible to keep the variance low. Later, in \cite{vargas2022unimodal}, the authors proposed using a beta distribution since it can achieve a small variance while keeping the mean at the centre of the class interval. In addition, they proposed a methodology to estimate the parameters of the distribution used to generate the label encoding for each class based on the total number of classes in the problem ($J$). They also conducted an extensive experimental study that showed that the beta distribution performs better than the previous alternatives on different datasets and considering different metrics. Finally, in \cite{vargas2023soft}, the authors proposed using triangular distributions instead of beta distributions. The most important benefit of triangular distributions is that they allow adjusting the error in adjacent classes by using a single parameter for the extreme classes and another one for the intermediate classes. The authors demonstrated through an experimental study that their proposal outperformed the alternatives proposed in \cite{liu2020unimodal} and achieved competitive results when compared with beta distributions proposed in \cite{vargas2022unimodal}.


Although the methods described in this paragraph obtain fairly good classification results when used to tackle ordinal classification problems, none of them allow for the prioritisation of extreme classes, which is desirable in certain problems as described above. Therefore, considering the importance of a good classification of the extreme classes and the excellent results obtained by the beta regularised loss function proposed in \cite{vargas2022unimodal}, the objectives of this work can be summarised as follows:
\begin{enumerate}
    \item To analyse the effect of employing a unimodal regularised loss function, based on a generalised beta distribution, for training a CNN model, examining its impact on the classification performance of the extreme classes.

    \item To propose a parameter estimation methodology for generalised beta distributions, which relies on restrictions on the mean and variance of the distributions, as well as the number of classes involved in the problem.

    \item To define a new evaluation metric to assess the classification performance specifically for the first and last classes.

    \item To test the methodology using six different ordinal benchmark datasets with different number of classes and five nominal and ordinal metrics, in addition to the proposed one.
    
\end{enumerate}

The rest of the manuscript is structured as follows: in \Cref{sec:methodology} the proposed regularisation based on a generalised beta distribution is presented, in \Cref{sec:parameters}, a method to estimate the parameters for the generalised beta distributions is proposed, \Cref{sec:experiments} describes the experimental design used to test the proposed methodology, in \Cref{sec:results} the results and the statistical analysis are shown, and, finally, in \Cref{sec:conclusions} the conclusions of this work are discussed.

\section{Methodology}
\label{sec:methodology}
This section describes the regularised loss function presented in this work, as well as the soft labelling baseline approach that forms the foundation for it.

\subsection{Baseline approach}
\label{sec:baseline}
The soft labelling approach introduced in \cite{liu2020unimodal} and later employed in several works including the one that proposed using beta distributions \cite{vargas2022unimodal} consists in replacing the loss function that is used to train a deep learning model with a regularised alternative where the standard labels are encoded with a soft alternative. To do that, they took the standard categorical cross-entropy loss function and defined its regularised alternative as follows:
\begin{equation}\label{eq:cce}
\mathscr{L}(\mathbf{x}, k) = \sum_{j=1}^{J} q'(j,k) [-\log P(y=C_j|\mathbf{x})],
\end{equation}
where $k = \classorder{k}$, $\class{k}$ is the target class and $q'(j,k) = P_j(k)$ defines the soft labels encoding. $P_j(k)$ is the probability for the $j$-th class when the actual target is the $k$-th category. 
This probability is directly given by the probability mass function (p.m.f.) of any given discrete distribution or can be obtained using the probability density function (p.d.f.) of any continuous distribution in the following manner:
\begin{equation}
    P_j(k) = \int\displaylimits_{(j-1)/J}^{j/J} f_k\left(x\right) \text{dx},
\end{equation}
where $f_k(x)$ is the p.d.f. or p.m.f. associated with class $\class{k}$. The aforementioned probability can be obtained using any type of probability distribution. However, an unimodal distribution that is centered in the interval of the true class should be the most appropriate for ordinal problems. Note that the probability is sampled between the interval limits associated with the $j$-th class. These intervals are defined by dividing the $[0,1]$ space in equal-length intervals based on the number of classes $J$ of the problem. Therefore, the lower and upper limits of the interval for class $\class{j}$ is given by $\frac{j-1}{J}$ and $\frac{j}{J}$, respectively. The definition of these intervals is based on the assumption that the probability distribution employed is bounded in the $[0,1]$ interval.

In the case of \cite{vargas2022unimodal}, beta distributions were employed to define the soft labels, and, therefore, the p.d.f. was expressed as follows:
\begin{equation}
\label{eq:beta}
    f(x; u, v) = \frac{x^{u-1}(1-x)^{v-1}}{B(u,v)}, \quad 0 < x < 1,
\end{equation}
where $u$ and $v$ are two parameters of the beta distribution that affect the location and the scale of it, and $B(u,v)$ is the Euler beta function, defined as:
\begin{equation}
    \euler{u,v} = \frac{\Gamma(u) \Gamma(v)}{\Gamma(u+v)}, \text{ where } \Gamma(u) = (u-1)!,
\end{equation}
for two positive integers $u$ and $v$.

\subsection{Proposed methodology}
\label{sec:betagen}
The regularised loss function based on generalised beta distributions, which is proposed in this work and described in this section, is an enhanced version of the methodology proposed in \cite{vargas2022unimodal}. It aims to improve the classification performance of the extreme classes by using a Generalised Beta (GB) distribution as a replacement for the standard beta distribution employed in that work. The generalised beta distribution can be denoted as $\text{GB}(\alpha, u, v)$ and adds a new parameter ($\alpha$) with respect to the standard beta distribution. The main goal of using a more flexible distribution is to improve the classification performance of the extreme classes. Said enhancement is achieved by reducing the variance of these distributions. In these terms, we consider using $\alpha = 2$ for the extreme classes and $\alpha = 1$ (standard beta) for the intermediate ones. Thus, the probability mass for the intermediate classes is concentrated in the middle of the interval while, for the first class it is around $x=0$, and for the last class most of the probability is around $x=1$.

In this way, the generalised beta distribution can be denoted as $\text{GB}(\alpha, u, v)$ and, a continuous random variable $X \sim \text{GB}(\alpha,u,v)$ has a p.d.f. $f(x;\alpha,u,v)$ defined by \cite{mcdonald1995generalization,mcdonald2008some} as:
\begin{equation}
\label{eq:gb}
    f(x; \alpha, u, v) = \frac{x^{\frac{u}{\alpha}-1} (1 - x^\frac{1}{\alpha})^{v-1}}{\alpha \euler{u,v}},
\end{equation}
where $\alpha > 0$, $u > 0$, $v > 0$ and B$(u,v)$ is the Euler beta function.

The existence of three free parameters makes this distribution very flexible and includes the standard beta distribution in the particular case of $\alpha=1$. Since the domain of $f(x;\alpha,u,v)$ is finite, all its moments are defined. The $h$-th order moment of a GB random variable is given by:
\begin{equation}
\label{eq:betamomenth}
    \expect{X^h} = \frac{\euler{u + \alpha h, v}}{\euler{u, v}}, \text{ for } u + \alpha h > 0.
\end{equation}
The demonstration of the process followed to achieve \Cref{eq:betamomenth} is shown in \ref{app:betamoments}. Then, the mean of the GB distribution is given by:
\begin{equation}
\label{eq:betamean}
\begin{aligned}
    \expect{X} &= \frac{\euler{u + \alpha, v}}{\euler{u, v}} =\\
    &= \frac{(u + \alpha - 1) \times \ldots \times u}{(u + v + \alpha -1) \times \ldots \times (u + v)}
    \text{ for } u + \alpha > 0,
\end{aligned}
\end{equation}
The second order moment is defined as:
\begin{equation}
\small
    \expect{X^2} = \frac{\euler{u + 2\alpha, v}}{\euler{u, v}} = \frac{(u + 2\alpha - 1) \times \ldots \times u}{(u + v + 2\alpha -1) \times \ldots \times (u + v)}, \text{ for } u + 2\alpha > 0.
\end{equation}
Then, the expression of the variance is determined by:
\begin{equation}
\small
    \variance{X} = \expect{X^2} - \expecttwo{X} = \frac{\euler{u + 2\alpha, v}}{\euler{u,v}} - \left(\frac{\euler{u + \alpha, v}}{\euler{u,v}}\right)^2, \text{ for } u + 2\alpha > 0,
\end{equation}
\begin{equation}
\footnotesize
    \variance{X} = \frac{(u +2\alpha -1) \times \ldots \times u}{(u+v+2\alpha-1) \times \ldots \times (u+v)} - \frac{(u+\alpha-1)^2 \times \ldots \times u^2}{(u+v+\alpha -1)^2 \times \ldots \times (u+v)^2}.
\end{equation}

In this way, the $P_j(k)$ term that was defined in the baseline approach described in \Cref{sec:baseline} can be defined according to the p.d.f. of the GB distribution:
\begin{equation}
    P_j(c) = \int\displaylimits_{(j-1)/J}^{j/J} f\left(x; \alpha_c, u_c, v_c\right) \text{dx},
\end{equation}
where $\alpha_c$, $u_c$ and $v_c$ are the parameters for the beta distribution associated to the true class $\class{c}$. These parameters, are calculated based on the true class and the number of categories of the problem in the same way described in \cite{vargas2022unimodal} for the intermediate classes, given that for these classes $\alpha=1$, which is the standard beta. However, the method employed for the first and the last classes is based on establishing some constraints for the mean and the variance of the distributions based on the target class, and it is described in \Cref{sec:parameters}. 

\section{Estimation of the parameters of the GB distribution as a function of the number of classes}
\label{sec:parameters}
Although there are different statistical methods for estimating the parameters of the GB$(\alpha, u, v)$ distribution, such as the method of the moments and the maximum likelihood method, they are too complex to be used to estimate the parameters of a generalised beta distribution due to the intractable integration expressions in the normalisation constant \citep{ma2010expectation}. For maximum likelihood estimation, numerical methods can be used to calculate the shape parameters of a generalised beta distribution using the smallest $M$ order statistics \citep{gnanadesikan1967maximum}. In \cite{narayanan1992note}, the authors show a numerically feasible method for parameter estimation in the multivariate beta (Dirichlet) distribution through the method of maximum likelihood. Also, in \cite{warsono2018estimation}, the authors use an iterative process to estimate the parameters of the generalised beta distribution. Finally, in \cite{makouei2021moments}, the authors derive recurrence relations for the single and the product moments of the order statistics as well as $k$-record values from the complementary beta distribution. However, the maximum likelihood method is known to provide poor results when the maximum is at the limit of the interval of one of the parameters. On the other hand, the method of moments is used to obtain a coarse approximation given that it gives an important bias when the samples sizes are not large enough.

In this context, new procedures for estimating the parameters of a GB distribution have been developed \citep{mano2021parameter,gomez2018family,kakamu2019bayesian}. However, they are not appropriate procedures for a soft-labelling procedure, given that we need to arrange the $J$ classes into $J$ sub-intervals are defined delimited by $J-2$ thresholds. Based on it, the probability for each sub-interval is used as a soft label. In this way, standard estimation procedures cannot be employed since the sample size and the composition of these samples in each of the sub-intervals are unknown.

\begin{figure}[!ht]
    \centering
    \includegraphics[width=\linewidth]{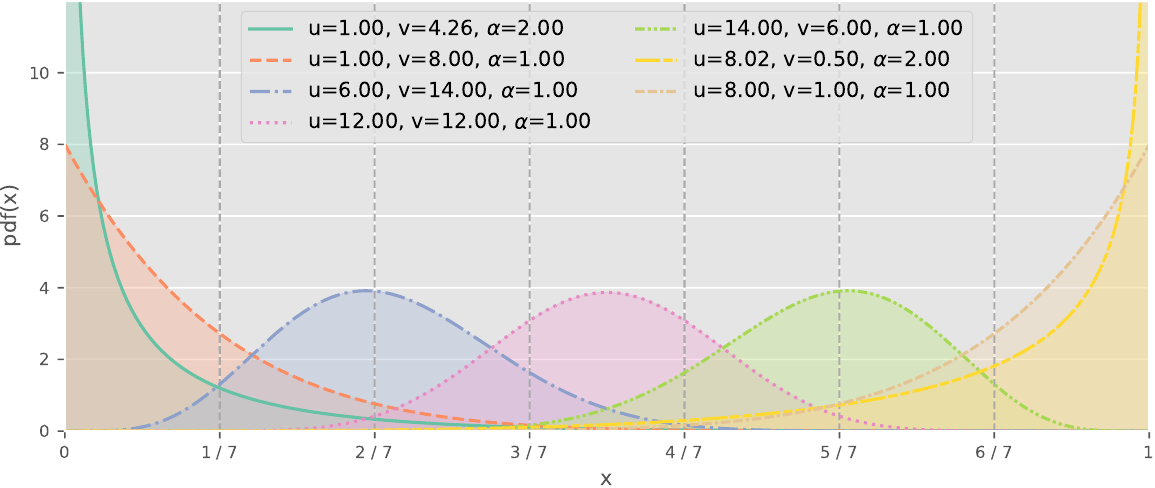}
    \caption{Probability density functions of the GB distributions for a problem with five classes. In the extreme classes, the standard beta ($\alpha=1$) is compared to the proposed alternative ($\alpha=2$).}
    \label{fig:gbplotextremes}
\end{figure}

It is worth noting that computing all the three parameters of the GB at the same time in a general form that can be used for any problem with a given number of classes is not a trivial task. Therefore, to simplify the process, the $\alpha$ parameter has been fixed while $u$ and $v$ have been computed using the method described in this section: i.e., they have been determined by constraining the values of $\expect{x}$ and $\variance{x}$ for each random variable associated with each projection sub-interval. In this way, the probability in the extremes is more concentrated around $0$ in the first class and around $1$ in the last class; while in the intermediate classes the probability should be concentrated around the midpoint of the sub-interval (see \Cref{fig:gbplotextremes}).

The parameter estimation procedure is composed of different steps. In a first step, the value of the $\alpha$ parameter is selected from $\{1.0, 2.0\}$. Based on these values, in a second step, values $\{0.5, 1.0\}$ are considered for the $u$ parameter of the first class, and values $\{0.5, 1.0\}$ are tested for the $v$ parameter of the last class. In a third step, taking into account the values of $\alpha$ and $u$, for the first class, or $v$, for the last class, the last parameter is obtained.

\Cref{fig:gbplotextremes} shows the distributions obtained for each target class for a problem with five classes. For the extreme classes, both the standard beta ($\alpha=1$) and the generalised beta ($\alpha=2$) are shown. As can be observed, when using $\alpha=2$, the means of the distributions of the extreme classes are closer to $0$, for the first class, and $1$, for the last class. Besides, this value reduces the variance of the distribution in comparison with $\alpha=1$, which is something desirable for extreme classes, as explained before. This fact is also justified in \ref{app:parameters}.

Thus, for the first class, the $v$ parameter is computed considering $\alpha=2$ and $u=1$, and, for the last class, the $u$ parameter is obtained using $\alpha=2$ and $v=0.5$. The decision of selecting these values is justified in \ref{app:parameters}. According to \Cref{eq:meanfirst,eq:varfirst}, the expression of the mean and the variance of the GB distribution when $\alpha=2$ and $u=1$ are given by:
\begin{equation}
    \expect{x} = \frac{2}{(v+2)(v+1)}, \text{ and } \variance{x} = \frac{44v^2 + 20v}{(v+4)(v+3)(v+2)^2(v+1)^2},
\end{equation}
which means that
\begin{equation}
    \lim\limits_{v \rightarrow \infty} \expect{x} = 0, \text{ and } \lim\limits_{v \rightarrow \infty} \variance{x} = 0.
\end{equation}
This is consistent with our hypothesis that the mean of the probability distribution of the first class is as close as possible to 0 and the variance is small.

On the other hand, for the last class, the parameters $\alpha = 2$ and $v=0.5$ are proposed. In this case, \Cref{eq:meanlast,eq:varlast} show that the mean of the distribution is given by:
\begin{equation}
    \expect{x} = \frac{4u(u+1)}{(2u+3)(2u+1)},
\end{equation}
and the variance is defined as:
\begin{equation}
    \variance{x} = \frac{128u^4+576u^3+736u^2+288u}{\left(2u+7\right)\left(2u+5\right)\left(2u+3\right)^2\left(2u+1\right)^2},
\end{equation}
so that
\begin{equation}
    \lim\limits_{u \rightarrow \infty} \expect{x} = 1, \text{ and } \lim\limits_{u \rightarrow \infty} \variance{x} = 0,
\end{equation}
which also matches the hypothesis that the probability distribution of the last class is as close as possible to 1 and the variance is close to 0. In contrast, for $v=1$,
\begin{equation}
    \lim\limits_{u \rightarrow \infty} \expect{x} = 1, \text{ but } \lim\limits_{u \rightarrow \infty} \variance{x} = 1
\end{equation}
instead of $0$, as show in \ref{app:parameters}. That is why using $v=0.5$ is more appropriate than using $v=1$ for the last class.

\subsection{Estimation of the parameters for the intermediate classes}
This section describes the methodology followed to obtain the parameters of the generalised beta distribution for the intermediate classes (from $2$ to $J-1$). Since the beta regularisation approach proposed in \cite{vargas2022unimodal} obtained very competitive results and the objective of this work is to improve the results in the extreme classes, for those problems where accurately classifying those classes is crucial, the same methodology described in that work is employed for the intermediate classes. Thus, the parameters for those classes are obtained by considering two constraints, which are based on the mean and the variance of the GB$(1, u, v)$ distributions:
\begin{equation}
    \frac{j}{J} \le \expect{X} \pm \stdev{X} \le \frac{j+1}{J}, \quad \text{ for } j=0,\ldots,J-1.
\end{equation}

\subsection{Estimation of the parameters for the first class}
Given that the $\alpha$ parameter is $2$, in order to obtain values of $v$ as a function of $u$ and the number of classes, two constraints associated with the mean and the variance of the distribution of the first class are proposed. Thus, the first expression constrains the mean of the distribution, so that it has to be between 0 and the centre of the first interval:
\begin{equation}
\label{eq:firstmean}
    \expect{X} = \frac{u(u+1)}{(u+v+1)(u+v)} \le \frac{1}{2J}.
\end{equation}

Then, the second constraint is associated with a linear combination of the mean and the standard deviation:
\begin{equation}
    0 \le \expect{X} - \lambda \stdev{X},~\stdev{X} \le \frac{1}{2J\lambda}, \text{ and } \variance{X} \le \frac{1}{4J^2\lambda^2},
\end{equation}
where $\lambda$ is the parameter that controls the linear combination and can be cross-validated for each dataset.

Given that $\variance{X} = \expect{X^2} - \expecttwo{X}$, the second constraint can be defined as follows:
\begin{equation}
    \expect{X^2} \le \frac{1}{4J^2\lambda^2} + \expecttwo{X} \le \frac{1 + \lambda^2}{4J^2\lambda^2}.
\end{equation}
Replacing the value of $\expecttwo{X}$ taking into account \Cref{eq:firstmean}:
\begin{equation}
\label{eq:firstmean2}
    \expect{X^2} = \frac{(u+3)(u+2)(u+1)u}{(u+v+3)(u+v+2)(u+v+1)(u+v)} \le \frac{1 + \lambda^2}{4J^2\lambda^2}.
\end{equation}
Then, replacing \Cref{eq:firstmean} in \Cref{eq:firstmean2}:
\begin{equation}
    \expect{X^2} = \frac{(u+3)(u+2)}{(u+v+3)(u+v+2)} \le \frac{1 + \lambda^2}{2J\lambda^2}.
\end{equation}

For simplicity, we define $F = \frac{1 + \lambda^2}{2J\lambda^2}$, so that $(u+3)(u+2) \le F(u+v+3)(u+v+2)$. Then, the following inequation defines a lower bound for the value of $v$:
\begin{equation}
    v \ge \frac{-F (2u+5) + \sqrt{F^2 (2u+5)^2 - 4F(F-1)(u^2+5u+6)}}{2F}.
\end{equation}
In this case, taking into account the values of $\alpha$ and $u$ which were previously selected for the first class ($\alpha=2$, $u=1$):
\begin{equation}
    v \ge \frac{-7F + \sqrt{F^2+48F}}{2F}.
\end{equation}
Using this expression, the lower bound of the $v$ parameter can be obtained based on the $\lambda$ parameter, which can be cross-validated.

\subsection{Estimation of the parameters for the last class}
The parameters of the last class are obtained in a similar way to those of the first class. Thus, considering the value $\alpha=2$, the first constraint can be set up as follows:
\begin{equation}
\label{eq:lastmean}
    \expect{X} = \frac{u(u+1)}{(u+v+1)(u+v)} = \frac{2J-1}{2J},
\end{equation}
while the second constraint can be defined as:
\begin{equation}
    \expect{X} + \eta \stdev{X} \le 1,
\end{equation}
and, consequently:
\begin{equation}
    \stdev{X} \le \frac{1 - \expect{X}}{\eta} \le \frac{1 - \frac{2J-1}{2J}}{\eta} = \frac{1}{2J\eta}.
\end{equation}
Then:
\begin{equation}
    \variance{X} = \expect{X^2} - \expecttwo{X} \le \frac{1}{4J^2\eta^2},
\end{equation}
and, replacing the value of $\expecttwo{X}$:
\begin{equation}
\label{eq:lastmean2}
    \expect{X^2} = \frac{(u+3)(u+2)(u+1)u}{(u+v+3)(u+v+2)(u+v+1)(u+v)} \le \frac{1 + \eta^2 (2J-1)^2}{4J^2\eta^2}.
\end{equation}
Using \Cref{eq:lastmean} and \Cref{eq:lastmean2}, we obtain:
\begin{equation}
    \expect{X^2} = \frac{(u+3)(u+2)}{(u+v+3)(u+v+2)} \le \frac{1 + \eta^2(2J-1)^2}{2J\eta^2(2J-1)}.
\end{equation}
Then, for simplicity, we define
\begin{equation}
    L = \frac{1 + \eta^2(2J-1)^2}{2J\eta^2(2J-1)},
\end{equation}
so that $(u+3)(u+2) \le L(u+v+3)(u+v+2)$. Solving the second degree inequation, an upper bound for $u$ is obtained:
\begin{equation}
    u \le \frac{-(5-6L) \pm \sqrt{(5-6L)^2 - 4(1-L)(6-(35/4)L)}}{2(1-L)}.
\end{equation}
This upper bound depends on the $\eta$ parameter, which can be cross-validated.

\section{Experiments}
\label{sec:experiments}
In this section, the experiments that were conducted to test the proposed method are described. In the first part of this section, the datasets considered are described. Then, the model is presented and, finally, the experimental design is explained.

\subsection{Datasets}
\label{sec:datasets}
In this section, several ordinal problems whose input data comes in the shape of images are described.

\subsubsection{Diabetic Retinopathy}
\begin{figure}[!ht]
    \centering
    \includegraphics[width=.19\linewidth]{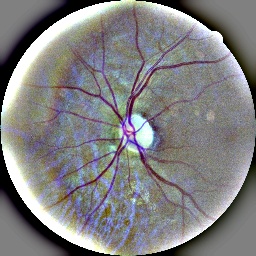}
    \includegraphics[width=.19\linewidth]{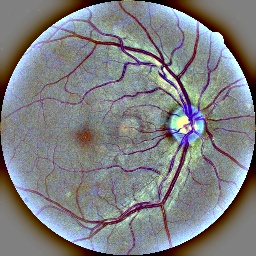}
    \includegraphics[width=.19\linewidth]{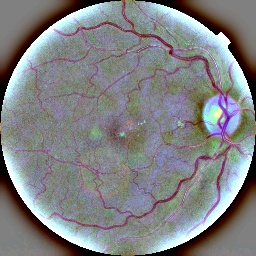}
    \includegraphics[width=.19\linewidth]{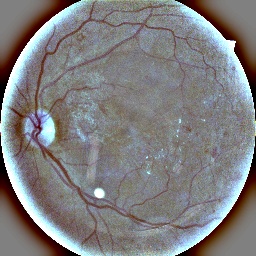}
    \includegraphics[width=.19\linewidth]{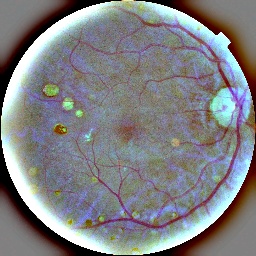}
    \caption{One image of each category ($0$ to $4$ from left to right) obtained from the training set of the diabetic retinopathy dataset.}
    \label{fig:retinopathy}
\end{figure}

Diabetic Retinopathy is a dataset of high-resolution eye fundus colour images. It was published in a Kaggle competition\footnote{\url{https://www.kaggle.com/c/diabetic-retinopathy-detection/data}} and since then has been commonly used as a benchmark dataset for ordinal classification methods \citep{wang2021automated,wu2020nfn,xie2020amd,de2020deep}. The dataset was provided as two separate splits for training and evaluation. The training set contains $17563$ pairs of images, from the left and right eyes. On the other hand, the testing set is composed of $26788$ pairs of images. Each image is labelled with one of five levels of diabetic retinopathy (DR) disease, where level 0 indicates a healthy patient while level 4 is associated with proliferative DR. The dataset is highly imbalanced given that most of the patients are healthy. Therefore, the number of samples on each of the categories is given by: $65342$ for level 0, $6205$ for level 1, $13152$ for level 2, $2087$ for level 3, and $1916$ for level 4. These images were taken with different devices and different lighting conditions. Therefore, to improve the training process and the generalisation capability, a preprocessing step, that was described in Kaggle\footnote{https://www.kaggle.com/ratthachat/aptos-eye-preprocessing-in-diabetic-retinopathy}, was performed aimed at enhancing the contrast of the images. Also, given that the original images are extremely high-resolution, they were resized to $256\times256$ to train the model. Some processed images from each category are shown in \Cref{fig:retinopathy}.

\subsubsection{Adience}
Adience \citep{eidinger2014age} is another benchmark dataset that is composed of colour images of human faces. Each image is associated with its gender and age. The age is given in different ranges, summing up to 8 classes. The whole dataset is composed of $26580$ faces belonging to $2284$ people. The images have been preprocessed in order to crop and align the faces, making the classification task easier. All the images have been resized to $256\times256$ and contrast-normalised. The original dataset was provided as five cross-validation folds. In this case, the first four folds are used for training and the last one for testing.

\subsubsection{FGNet}
FGNet \citep{fu2014interestingness} is a smaller faces images dataset composed of $1002$ colour images with $128\times128$ resolution. These images belong to $82$ different subjects. Each of the samples is labelled with the exact age of the person in the moment when the photograph was taken. Therefore, different categories can be obtained depending on the age grouping established. For this work, we have grouped the samples using the following intervals: $[0, 3)$, $[3,11)$, $[11,16)$, $[16,24)$, $[24,40)$, $[40,+\infty)$. Given that the whole dataset was provided without partitions, $20\%$ of the data was used for testing and the rest for training. These partitions were created in a stratified way, keeping the same ratio of samples from each class than in the original dataset.

\subsubsection{UTKFace}
UTKFace \citep{zhang2017age} is a face dataset with a long age span (ranging from 0 to 116 years old). The dataset is composed of $20,000$ face images with annotations of age, gender and ethnicity. These images cover large variations in pose, facial expression, illumination conditions, occlusion, resolution, etc. The dataset can be used for different types of tasks like face detection, age estimation or landmark localisation. In this case, we are interested in solving the problem related to age estimation, given that it is an ordinal task. The original images are already cropped and aligned and, thus, on each image, there is only one visible face. Every image is provided along with its corresponding meta-data: an integer indicating the exact age, the gender as a binary variable, the race as an integer, and the date and time when the picture was collected. In this work, we are only using the age variable. In order to convert the continuous age to an ordinal variable, $12$ categories have been determined using the following intervals: $[0, 2)$, $[2, 6)$, $[6, 12)$, $[12, 19)$, $[19, 23)$, $[23, 27)$, $[27, 30)$, $[30, 38)$, $[38, 45)$, $[45, 55)$, $[55, 65)$, $[65, 73)$, $[73, 80)$, $[80, +\infty)$. Then, the training and test partitions were created taking $80\%$ and $20\%$ of the complete set, respectively, in a stratified way.

\subsubsection{Aesthetic Visual Analysis}
\begin{figure}[!ht]
    \centering
    \includegraphics[width=.15\linewidth]{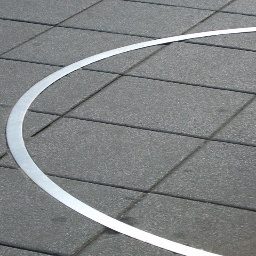}
    \includegraphics[width=.15\linewidth]{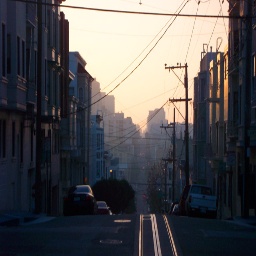}
    \includegraphics[width=.15\linewidth]{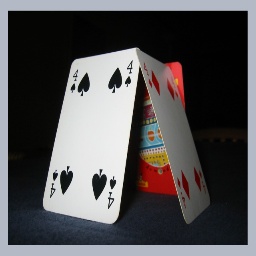}
    \includegraphics[width=.15\linewidth]{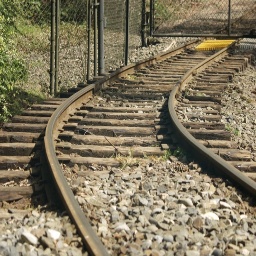}
    \includegraphics[width=.15\linewidth]{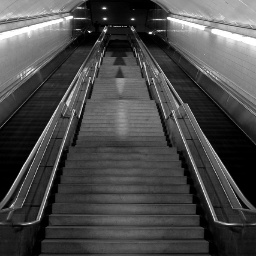}
    \includegraphics[width=.15\linewidth]{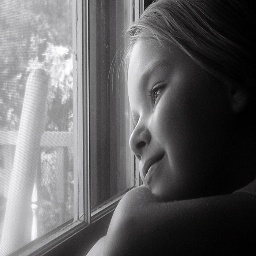}
    \caption{Images obtained from the AVA dataset.}
    \label{fig:ava}
\end{figure}

The Aesthetic Visual Analysis (AVA) dataset was introduced in 2012 as a new benchmark dataset \citep{murray2012ava}. AVA contains over $250,000$ images along with a rich variety of meta-data, including a large number of aesthetic scores for each image, semantic labels for over 60 categories as well as labels related to the photographic style. In this work, our interest lies into the aesthetic scores, as they can be grouped into different intervals and be used as ordinal labels. Each image received from $78$ to $549$ votes in a range of $[0,10]$ and the information is provided as the number of votes on each of the ratings. To convert these individual ratings to an ordinal label, the mean value was computed. Then, the final ordinal label was assigned based on a series of subsets: $[0,2]$, $\{3\}$, $\{4\}$, $\{5\}$, $\{6\}$, and $[7,10]$. The original images come with different resolutions. Thus, to work with them, they have been rescaled to $256\times256$. Given that the original set was provided without any partition, the test size was constructed taking a stratified random sample containing $20\%$ of the data. The rest was taken for the training set. \Cref{fig:ava} represents some of the images of the AVA dataset. Those images were taken from different categories, where the first belongs to the lowest rating and the last to the highest one.

\subsubsection{WIKI (IMDB-WIKI)}
The IMDB-WIKI \citep{rothe2018deep} dataset is one of the largest public datasets for age prediction. It contains $500,000$ images obtained from IMDb and Wikipedia. In this work, we use a small subset of this dataset that was obtained using only the images from Wikipedia. Therefore, the total number of images is reduced to $62,328$. From these images, $80\%$ are used for training the model, and the remaining $20\%$ for evaluation. Also, this dataset requires a preprocessing step before using these images to train a model because some of them are blank, too low resolution or do not contain any faces. Thus, in the preprocessing step, we only selected those images that have are not blank and have a resolution higher than $100\times100$. Also, a face detection algorithm was employed to find the bounding box of the face and obtain a cropped and aligned image. After that, all the images were resized to $128\times128$.

\begin{figure}[!ht]
    \centering
    \includegraphics[width=.15\linewidth]{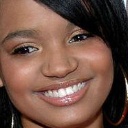}
    \includegraphics[width=.15\linewidth]{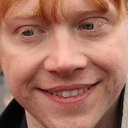}
    \includegraphics[width=.15\linewidth]{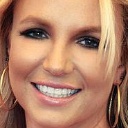}
    \includegraphics[width=.15\linewidth]{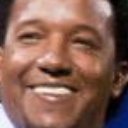}
    \includegraphics[width=.15\linewidth]{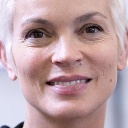}
    \includegraphics[width=.15\linewidth]{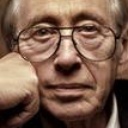}
    \caption{Images extracted from the different categories of the WIKI dataset.}
    \label{fig:wiki}
\end{figure}

All the images are provided along with the date of birth of the subject and the date when the picture was taken. In this way, the age of the person at the moment the photo was taken can be calculated. After that, $6$ age ranges are established to define the ordinal categories: $[0,24)$, $[24,29)$, $[29,34)$, $[34,45)$, $[45,55)$, $[55,+\infty)$. Those intervals have been defined taking into account the number of existing samples on each age range. The first category ($[0,24)$) is wider than the other given that there are few samples with an early age. The same thing occurs with the last category. Some examples from this dataset can be observed in \Cref{fig:wiki}.

\subsection{Model}
\label{sec:model}
Convolutional Neural Networks (CNN) are the most popular alternative to work with image datasets. In this case, we are using a well-known architecture that has achieved very competitive performance with a fairly small number of parameters. This model is commonly named ResNet18, and is the smallest type of residual network that has been widely used in previous works \citep{he2016deep}. The main advantage of using an already existing model is the possibility of considering pre-trained weights. In this case, the PyTorch library was used, and the model was initialised using the ImageNet weights. This kind of initialisation significantly enhances the convergence speed, thus, reducing the training time by decreasing the required number of training epochs. The output layer was replaced with a fully connected layer with $J$ units, being $J$ the number of classes of each dataset. The weights of this layer were randomly initialised given that the pre-trained weights are based on a fully connected layer for a problem with $1000$ classes (like ImageNet). The activation function at the output of the model is the standard softmax, which outputs probabilities for each of the categories.

\subsection{Experimental design}
\label{sec:experimentaldesign}
The model described in \Cref{sec:model} was trained and evaluated using the process described in this Section. First of all, $15\%$ of the training set was split for validation in a stratified way. During the training process, the data was fed into the model using batches of $200$ elements. The batch size was adjusted to find a balance between the computational time spent for the training process and the memory required to fit all the data into the GPUs memory. The same batch size was kept for all the datasets.

The Adam algorithm was selected for the optimisation process \citep{kingma2014adam}, as it has been proved to obtain good performance in previous works \citep{na2022efficient,arcos2018deep}. The learning rate for the optimiser has been adjusted every 7 epochs, multiplying it by a factor of $0.5$. The initial learning rate was cross-validated using the training set and fixed to $10^{-3}$ for all the datasets. Also, the number of training epochs was adjusted in order to maximise the performance and minimise the computational time. Doing extra optimisation steps does not harm the performance, given that, at the end of the training process, the best model weights are selected taking into account the validation metrics. However, doing extra unneeded optimisation epochs can significantly increase the training time. In this way, the training epochs number was fixed to $25$. It is worth noting that using the aforementioned pre-trained weights helps significantly reduce the required number of training epochs. 

For the loss function considered during the optimisation process, different alternatives were tested:
\begin{itemize}
    \item Standard beta regularised CCE, proposed in \cite{vargas2022unimodal}.
    \item Triangular regularised CCE, proposed in \cite{vargas2023soft}.
    \item Generalised beta regularised CCE. This is the loss function proposed in this work.
\end{itemize}

In the case of the proposed methodology, the parameters $\lambda$ and $\eta$, which are part of the constraints considered to calculate the parameters of the distributions used for the first and the last class, are cross-validated using the validation set in such a way that $\lambda, \eta \in \{0.5, 0.75, 1.0, 1.25, 1.5\}$.

For each of the loss types and datasets, $30$ executions were performed. For each execution, a new hold-out for training and validation was obtained using the selected random seed. Also, the initialisation of the parameters of the fully-connected layers is done based on the random seed. The convolutional blocks are initialised using the ImageNet pre-trained weights.

\subsection{Performance metrics}
\label{sec:metrics}
With the aim to compare the proposed methodology against the previous alternatives which were proposed in the literature, six performance metrics are considered. Some of the are standard classification metrics (Accuracy), some others are intended for ordinal classification (WK, MAE and 1-off), and the other two are well-suited for unbalanced problems (MS and GMSEC). Specifically, the proposed GMSEC evaluates the performance in the extreme classes.

\begin{itemize}
    \item Weighted Kappa (WK) \citep{de2018weighted}, which is based on the standard Kappa ($\kappa$) index. In this case, quadratic weights are used. This metric can be defined as follows:
    \begin{equation}
        \kappa_w = 1 - \frac{\sum\limits^N_{ij} \omega_{ij} O_{ij}}{\sum\limits^N_{ij} \omega_{ij} E_{ij}},
    \end{equation}
    where $N$ is the number of samples, $\omega_{ij}$ are the elements of the penalisation matrix (in this case, quadratic weights are considered, $\omega_{ij} = \frac{(i-j)^2}{(J-1)^2}$, $\omega_{ij} \in [0,1]$), $O_{ij}$ are the elements of the confusion matrix, $E_{ij} = \frac{O_{i\bullet} O_{\bullet j}}{N}$, $O_{i\bullet}$ is the sum of the elements of the $i\text{-th}$ row, and $O_{\bullet j}$ is the sum of the elements of the $j\text{-th}$ column.
    
    \item Minimum Sensitivity (MS) \citep{caballero2010sensitivity,cruz2014metrics}, which is computed as the minimum value of sensitivity computed for each class. Thus, this metric represents the lowest percentage of patterns correctly classified as belonging to each class, concerning the total number of samples on the class:
    \begin{equation}
        \text{MS} = \min\Big\{S_j = \frac{O_{jj}}{O_{j\bullet}}; j = 1, ..., J\Big\},
    \end{equation}
    where $S_j$ is the sensitivity of class $j$.
    Using this metric, we ensure that all of the classes of each problem are decently classified.
    
    \item Mean Absolute Error (MAE) \citep{cruz2014metrics}. The MAE is the average absolute deviation of the predicted class from the correct class. It can be computed as follows:
    \begin{equation}
        \text{MAE} = \frac{1}{N} \sum_{i=1}^J \sum_{j=1}^J |i-j|O_{ij}.
    \end{equation}
    
    \item Accuracy or correct classification rate (CCR).
    
    \item 1-off accuracy, which accounts a sample as correctly classified when it is predicted in the category correct or in an adjacent class according to the ordinal scale.

    \item Geometric Mean of the Sensitivity of Extreme Classes (GMSEC). This new proposal is defined as a metric used to measure the quality of the classification for the extreme categories of any given problem. This metric is based on the geometric mean of the individual sensitivities of the first and the last classes in the ordinal scale. Using this metric is important for those problems where the first and the last categories are more important than the intermediate ones because of the intrinsic characteristics of the real problem. Thus, the GMSEC can be defined as follows:
    \begin{equation}
        \text{GMSEC} = \sqrt{S_1 \cdot S_J},
    \end{equation}
    which can be also expressed in the following manner:
    \begin{equation}
        \text{GMSEC} = e^{\frac{1}{2} (\ln S_1 + \ln S_J)}.
    \end{equation}
\end{itemize}

\section{Results}
\label{sec:results}
In this section, the results of the experiments are described taking into account the performance metrics described in \Cref{sec:metrics}. Thus, in \Cref{table:results}, the averaged results of $30$ executions of each dataset and loss function are shown.

\begin{table}[!ht]
\centering
\scriptsize
\begin{tabular}{lcccccc}
\toprule
Loss & QWK \textuparrow & MS \textuparrow & MAE \textdownarrow & CCR \textuparrow & 1-off \textuparrow & GMSEC \textuparrow \\
\midrule
\multicolumn{7}{c}{Adience}\\
\midrule
CCE-$\beta_{gen}$ & \bfseries 0.864\textsubscript{0.005} & \itshape 0.258\textsubscript{0.022} & \bfseries 0.675\textsubscript{0.012} & \itshape 0.479\textsubscript{0.008} & \bfseries 0.892\textsubscript{0.005} & \bfseries 0.497\textsubscript{0.041} \\
CCE-$\beta$ & \itshape 0.863\textsubscript{0.004} & \bfseries 0.265\textsubscript{0.026} & \itshape 0.676\textsubscript{0.008} & \itshape 0.479\textsubscript{0.007} & \bfseries 0.892\textsubscript{0.004} & 0.474\textsubscript{0.023} \\
CCE-T & 0.858\textsubscript{0.006} & 0.232\textsubscript{0.035} & \itshape 0.676\textsubscript{0.013} & \bfseries 0.490\textsubscript{0.013} & 0.884\textsubscript{0.008} & \itshape 0.483\textsubscript{0.037} \\ \midrule
\multicolumn{7}{c}{AVA}\\
\midrule
CCE-$\beta_{gen}$ & \bfseries 0.348\textsubscript{0.013} & \bfseries 0.158\textsubscript{0.031} & \itshape 0.853\textsubscript{0.028} & \itshape 0.349\textsubscript{0.010} & \itshape 0.822\textsubscript{0.014} & \itshape 0.245\textsubscript{0.040} \\
CCE-$\beta$ & 0.335\textsubscript{0.017} & \itshape 0.141\textsubscript{0.034} & 0.895\textsubscript{0.065} & 0.334\textsubscript{0.023} & 0.802\textsubscript{0.032} & \bfseries 0.292\textsubscript{0.048} \\
CCE-T & \itshape 0.345\textsubscript{0.012} & \itshape 0.141\textsubscript{0.041} & \bfseries 0.805\textsubscript{0.069} & \bfseries 0.372\textsubscript{0.032} & \bfseries 0.844\textsubscript{0.031} & 0.231\textsubscript{0.058} \\ \midrule
\multicolumn{7}{c}{FGNet}\\
\midrule
CCE-$\beta_{gen}$ & \itshape 0.861\textsubscript{0.014} & \itshape 0.376\textsubscript{0.043} & 0.503\textsubscript{0.032} & 0.557\textsubscript{0.029} & \itshape 0.949\textsubscript{0.010} & \bfseries 0.670\textsubscript{0.072} \\
CCE-$\beta$ & \bfseries 0.862\textsubscript{0.015} & \bfseries 0.387\textsubscript{0.058} & \bfseries 0.498\textsubscript{0.038} & \itshape 0.561\textsubscript{0.035} & \bfseries 0.952\textsubscript{0.013} & \itshape 0.665\textsubscript{0.084} \\
CCE-T & 0.858\textsubscript{0.017} & 0.364\textsubscript{0.057} & \bfseries 0.498\textsubscript{0.041} & \bfseries 0.571\textsubscript{0.031} & 0.944\textsubscript{0.015} & 0.653\textsubscript{0.074} \\ \midrule
\multicolumn{7}{c}{Retinopathy}\\
\midrule
CCE-$\beta_{gen}$ & \bfseries 0.618\textsubscript{0.007} & \itshape 0.066\textsubscript{0.036} & \itshape 0.437\textsubscript{0.017} & \itshape 0.684\textsubscript{0.018} & \itshape 0.906\textsubscript{0.006} & \itshape 0.560\textsubscript{0.040} \\
CCE-$\beta$ & 0.442\textsubscript{0.037} & 0.022\textsubscript{0.030} & 0.872\textsubscript{0.137} & 0.198\textsubscript{0.148} & \bfseries 0.954\textsubscript{0.014} & 0.148\textsubscript{0.147} \\
CCE-T & \itshape 0.614\textsubscript{0.007} & \bfseries 0.137\textsubscript{0.056} & \bfseries 0.422\textsubscript{0.017} & \bfseries 0.700\textsubscript{0.021} & 0.898\textsubscript{0.008} & \bfseries 0.570\textsubscript{0.026} \\ \midrule
\multicolumn{7}{c}{UTKFace}\\
\midrule
CCE-$\beta_{gen}$ &\bfseries 0.753\textsubscript{0.006} & \itshape 0.164\textsubscript{0.041} & 1.024\textsubscript{0.017} & \itshape 0.474\textsubscript{0.006} & \bfseries 0.840\textsubscript{0.005} & \bfseries 0.591\textsubscript{0.057} \\
CCE-$\beta$ & \bfseries 0.753\textsubscript{0.006} & 0.152\textsubscript{0.046} & \bfseries 1.022\textsubscript{0.015} & 0.473\textsubscript{0.006} & \bfseries 0.840\textsubscript{0.005} & 0.570\textsubscript{0.057} \\
CCE-T & \bfseries 0.753\textsubscript{0.010} & \bfseries 0.173\textsubscript{0.047} & \itshape 1.023\textsubscript{0.025} & \bfseries 0.475\textsubscript{0.007} & \bfseries 0.840\textsubscript{0.007} & \itshape 0.580\textsubscript{0.061} \\ \midrule
\multicolumn{7}{c}{Wiki}\\
\midrule
CCE-$\beta_{gen}$ & \bfseries 0.797\textsubscript{0.004} & \bfseries 0.339\textsubscript{0.032} & \bfseries 0.714\textsubscript{0.009} & \bfseries 0.456\textsubscript{0.005} & \itshape 0.866\textsubscript{0.005} & \bfseries 0.507\textsubscript{0.039} \\
CCE-$\beta$ & 0.794\textsubscript{0.004} & 0.311\textsubscript{0.026} & \itshape 0.715\textsubscript{0.009} & 0.452\textsubscript{0.006} & \bfseries 0.869\textsubscript{0.005} & 0.462\textsubscript{0.038} \\
CCE-T & \itshape 0.795\textsubscript{0.003} & \itshape 0.332\textsubscript{0.027} & 0.716\textsubscript{0.008} & \itshape 0.455\textsubscript{0.004} & 0.865\textsubscript{0.005} & \itshape 0.500\textsubscript{0.033} \\
\bottomrule
\end{tabular}
\caption{Mean results of 30 executions for each dataset and each metric. The values in the table show the mean value along with the standard deviation (Mean\textsubscript{SD}). Note that the best value in each column for each dataset is highlighted in bold and the second best in italics.}
\label{table:results}
\end{table}

\subsection{Statistical analysis}
A statistical analysis has been performed to compare the different loss functions. First, we set the hypothesis that the results, for each metric, and considering the 30 runs for all the datasets and methodologies, follow a normal distribution. To do this, a Kolmogorov-Smirnov test was performed, having $p$-values higher, in all cases, than the level of significance ($\alpha=0.05$). Therefore, the results confirm that the values of all the metrics are normally distributed. However, taking into account the experimental design, the results are dependent, so an analysis of variance test of two factors (ANOVA II) is not appropriate for this study.

Thus, each pair of methodologies was compared using a Student's t-test. The level of significance ($\alpha$) was set to $0.05$. Given that multiple comparisons were made, the corresponding correction associated to the number of comparisons was applied. As $3$ algorithms are compared, the total number of comparisons for each dataset is $3$. Therefore, the corrected level of significance is $\alpha^* = 0.05/3 = 0.016$. The total number of statistically significant wins (W), draws (D) and losses (L) are shown in \Cref{table:student}. For each metric of that table, the highest number of wins and the lowest number of losses are highlighted using bold font on each column.

\begin{table}[!ht]
\centering
\begin{tabular}{l@{\hskip 0.25cm}*{18}{c@{\hskip 0.25cm}}}
\toprule
    & \multicolumn{3}{c}{QWK} & \multicolumn{3}{c}{MAE} & \multicolumn{3}{c}{CCR} & \multicolumn{3}{c}{1-off} & \multicolumn{3}{c}{MS} & \multicolumn{3}{c}{GMSEC}\\
    \midrule
    Method & W & D & L & W & D & L & W & D & L & W & D & L & W & D & L & W & D & L\\
    \midrule
    CCE-$\beta_{gen}$ & $\mathbf{5}$ & $\mathbf{7}$ & $\mathbf{0}$ & $2$ & $8$ & $2$ & $3$ & $6$ & $3$ & $3$ & $6$ & $3$ & $\mathbf{4}$ & $\mathbf{7}$ & $\mathbf{1}$ & $\mathbf{4}$ & $\mathbf{7}$ & $\mathbf{1}$\\
    
    CCE-$\beta$ & $1$ & $6$ & $5$ & $0$ & $8$ & $4$ & $0$ & $5$ & $7$ & $\mathbf{6}$ & $\mathbf{4}$ & $\mathbf{2}$ & $1$ & $6$ & $5$ & $2$ & $5$ & $5$\\
    
    CCE-T & $2$ & $7$ & $3$ & $\mathbf{4}$ & $\mathbf{8}$ & $\mathbf{0}$ & $\mathbf{7}$ & $\mathbf{5}$ & $\mathbf{0}$ & $2$ & $4$ & $6$ & $3$ & $7$ & $2$ & $2$ & $8$ & $2$\\
    
    \bottomrule
\end{tabular}
\caption{Student's t-test results over all datasets and metrics.}
\label{table:student}
\end{table}

The results of the t-test represented in \Cref{table:student} show that the methodology that uses generalised beta distribution to define the soft labels is the one with the best result for the GMSEC metric, which describes the performance in the extreme classes. In addition, it also obtained the best results for QWK and MS metrics, maintaining competitive results in MAE, CCR and 1-off metrics.

In this way, the proposed methodology improves the sensitivities of the extreme classes without a significant detriment of the sensitivities of the intermediate classes for the six datasets considered in the experimental design.

\section{Conclusions}
\label{sec:conclusions}
In this work, a new unimodal regularisation approach for the loss function based on a generalised beta distribution was proposed. Also, a method to obtain the optimal distribution parameters for each class for a problem with any number of classes was presented. The main advantage of the proposed distribution concerning the standard beta distribution lies on the enhancement of the classification performance of the extreme classes, while keeping a good classification performance for the rest of the categories. The proposed approach was tested using 6 different datasets with variable number of classes. The average results of 30 executions for each dataset were compared with the results of the standard beta and other unimodal distributions proposed in previous works. The experimental results and the posterior statistical analysis showed that the proposed approach performed better than the other compared alternatives. Also, a more in-depth analysis was performed taking into account the standard beta and the proposed generalised beta to check whether the latter improved the sensitivity of the extreme classes. To do that, the GMSEC metric was introduced and the results shown that, effectively, the proposed distribution improved the classification performance for the extreme classes.

\section*{Acknowledgments}
This work has been partially subsidised by ``Agencia Española de Investigaci\'on (España)'' (grant ref.: PID2020-115454GB-C22 / AEI / 10.13039 / 501100011033). V\'ictor Manuel Vargas's research has been subsidised by the FPU Predoctoral Program of the Spanish Ministry of Science, Innovation and Universities (MCIU), grant reference FPU18/00358.

\bibliographystyle{model5-names}
\bibliography{main} 

\appendix

\section{GB moments demonstration}
\label{app:betamoments}
Considering the definition of the p.d.f. of the GB distribution provided in \Cref{eq:gb}, the $h$-th-order moment of this distribution can be defined as follows:
\begin{equation}
    \expect{X} = \frac{B(\alpha h + u, v)}{B(u,v)}, \text{ for } u + \alpha h > 0.
\end{equation}

\subsection{Demonstration}
To prove it, we change $x^\frac{1}{\alpha}$ for $z$ so that:
\begin{equation}
    x = z^{\alpha}, \text{ and } \diff x = \alpha z^{\alpha - 1} \diff z,
\end{equation}
so:
\begin{equation}
    \expect{X^h} = \frac{1}{\alpha B(u,v)} \int_0^1 x^{h+\frac{u}{\alpha}-1}(1-x^\frac{1}{\alpha})^{v-1} \diff x,
\end{equation}
and reverting the change of variables of $z$, we obtain:
\begin{equation}
\small
    \begin{aligned}
        \expect{X^h} &= \frac{1}{\alpha B(u,v)} \int_0^1 z^{\alpha h + u-\alpha} (1-z)^{v-1} \alpha z^{\alpha - 1} \diff z =\\
        &= \frac{1}{\euler{u,v}} \int_0^1 z^{\alpha h + u -1} (1 - z)^{v-1} \diff z =\frac{B(\alpha h + u, v)}{B(u,v)}, \quad u + \alpha h > 0.
    \end{aligned}
\end{equation}

\section{GB parameters selection}
\label{app:parameters}
The mean of the distribution for $\alpha=2$ is given by:
\footnotesize
\begin{align}
    &\expect{x} = \frac{\euler{u+2, v}}{\euler{u, v}} = \frac{u(u+1)}{(u+v+1)(u+v)}.\\
    \label{eq:meanfirst}&\text{If } u=1,~ \expect{x} = \frac{2}{(v+2)(v+1)}; \text{ if } v \rightarrow \infty,~ \expect{x} \rightarrow 0.\\
    &\text{If } v = 1,~ \expect{x} = \frac{u}{u+2};~ \text{if } u \rightarrow \infty, ~ \expect{x} \rightarrow 1.\\
    \label{eq:meanlast}&\text{If } v = 0.5, ~\expect{x} = \frac{4u(u+1)}{(2u+3)(2u+1)}; \text{ if } u \rightarrow \infty, ~\expect{x} \rightarrow 1.
\end{align}
\normalsize

On the other hand, the mean of the distribution for $\alpha=1$ is obtained as follows:
\footnotesize
\begin{align}
    & ~ \expect{x} = \frac{\euler{u + 1, v}}{\euler{u, v}} = \frac{u}{u+v}.\\
    &\text{If } u=1, ~ \expect{x} = \frac{1}{v + 1}; \text{ if } v \rightarrow \infty, ~ \expect{x} \rightarrow 0.\\
    &\text{If } v=1, ~ \expect{x} = \frac{u}{u+1}; \text{ if } u \rightarrow \infty, ~ \expect{x} \rightarrow 1.
\end{align}
\normalsize

Then, the variance of the distribution for $\alpha=2$ is computed in the following manner:
\footnotesize
\begin{align}
    &\begin{aligned}
    &\variance{x} = \frac{\euler{u+4, v}}{\euler{u,v}} - \left( \frac{\euler{u+2, v}}{\euler{u,v}} \right)^2 =\\ &=\frac{u(u+3)(u+2)(u+1)}{(u+v+3)(u+v+2)(u+v+1)(u+v)} - \left( \frac{u(u+1)}{(u+v+1)(u+v)} \right)^2.
    \end{aligned}\\
    \label{eq:varfirst}&\text{If } u=1, \variance{x} = \frac{20v + 44v^2}{(v+4)(v+3)(v+2)^2(v+1)^2}; \text{ if } v \rightarrow \infty, ~ \variance{x} \rightarrow 0.\\
    &\text{If } v = 1, \variance{x} = \frac{u^3 + 3u^2 + 4u}{(u+4)(u+2)^2}; \text{ if } u \rightarrow \infty, ~ \variance{x} \rightarrow 1.\\
    \label{eq:varlast}&\text{If } v=0.5, \variance{x} = \frac{128u^4+576u^3+736u^2+288u}{\left(2u+7\right)\left(2u+5\right)\left(2u+3\right)^2\left(2u+1\right)^2}; \text{ if } u \rightarrow \infty, ~ \variance{x} \rightarrow 0.
\end{align}
\normalsize

Finally, the variance of the distribution for $\alpha=1$ is given by:
\footnotesize
\begin{align}
&\variance{x} = \frac{\euler{u+2,v}}{\euler{u,v}} - \left( \frac{\euler{u+1,v}}{\euler{u,v}} \right)^2 = \frac{uv}{(u+v+1)(u+v)^2}.\\
&\text{If } u=1, \variance{x} = \frac{v}{(v+2)(v+1)^2}; \text{ if } v \rightarrow \infty, ~ \variance{x} \rightarrow 0.\\
&\text{If } v=1, \variance{x} = \frac{u}{(u+2)(u+1)^2}; \text{ if } u \rightarrow \infty, ~ \variance{x} \rightarrow 0.\\
&\text{If } v=0.5,~\variance{x} = \frac{4u}{(2u+3)(2u+1)^2}; \text{ if } u \rightarrow \infty,~\variance{x} \rightarrow 0.
\end{align}
\normalsize

\end{document}